\pdfoutput=1

\documentclass[11pt]{article}

\usepackage[final]{acl}

\usepackage{times}
\usepackage{latexsym}
\usepackage{amsmath}
\usepackage{algorithm}
\usepackage{algorithmic}   

\usepackage[T1]{fontenc}

\usepackage[utf8]{inputenc}

\usepackage{microtype}

\usepackage{inconsolata}

\usepackage{graphicx}

\title{Beyond Generation: Multi-Hop Reasoning for Factual Accuracy in Vision-Language Models}

\author{
  Shamima Hossain$^{1,2}$ \\
  $^{1}$Department of Computer Science, Brac University, Bangladesh\\
  $^{2}$bKash Limited, Dhaka, Bangladesh\\
  \texttt{shamima.hossain@g.bracu.ac.bd, shamima.alma@bkash.com}
}

\begin{document}
\maketitle
\begin{abstract}
Visual Language Models (VLMs) are powerful generative tools but often produce factually inaccurate outputs due to a lack of robust reasoning capabilities. While extensive research has been conducted on integrating external knowledge for reasoning in large language models (LLMs), such efforts remain underexplored in VLMs, where the challenge is compounded by the need to bridge multiple modalities seamlessly. This work introduces a framework for knowledge-guided reasoning in VLMs, leveraging structured knowledge graphs for multi-hop verification using image-captioning task to illustrate our framework. Our approach enables systematic reasoning across multiple steps, including visual entity recognition, knowledge graph traversal, and fact-based caption refinement. We evaluate the framework using hierarchical, triple-based and bullet-point based knowledge representations, analyzing their effectiveness in factual accuracy and logical inference. Empirical results show that our approach improves factual accuracy by approximately 31\% on preliminary experiments on a curated dataset of mixtures from Google Landmarks v2, Conceptual captions and Coco captions revealing key insights into reasoning patterns and failure modes. This work demonstrates the potential of integrating external knowledge for advancing reasoning in VLMs, paving the way for more reliable and knowledgable multimodal systems.
\end{abstract}

\section{Introduction}
\begin{figure}[t]
\hfill
  \includegraphics[width=\linewidth]{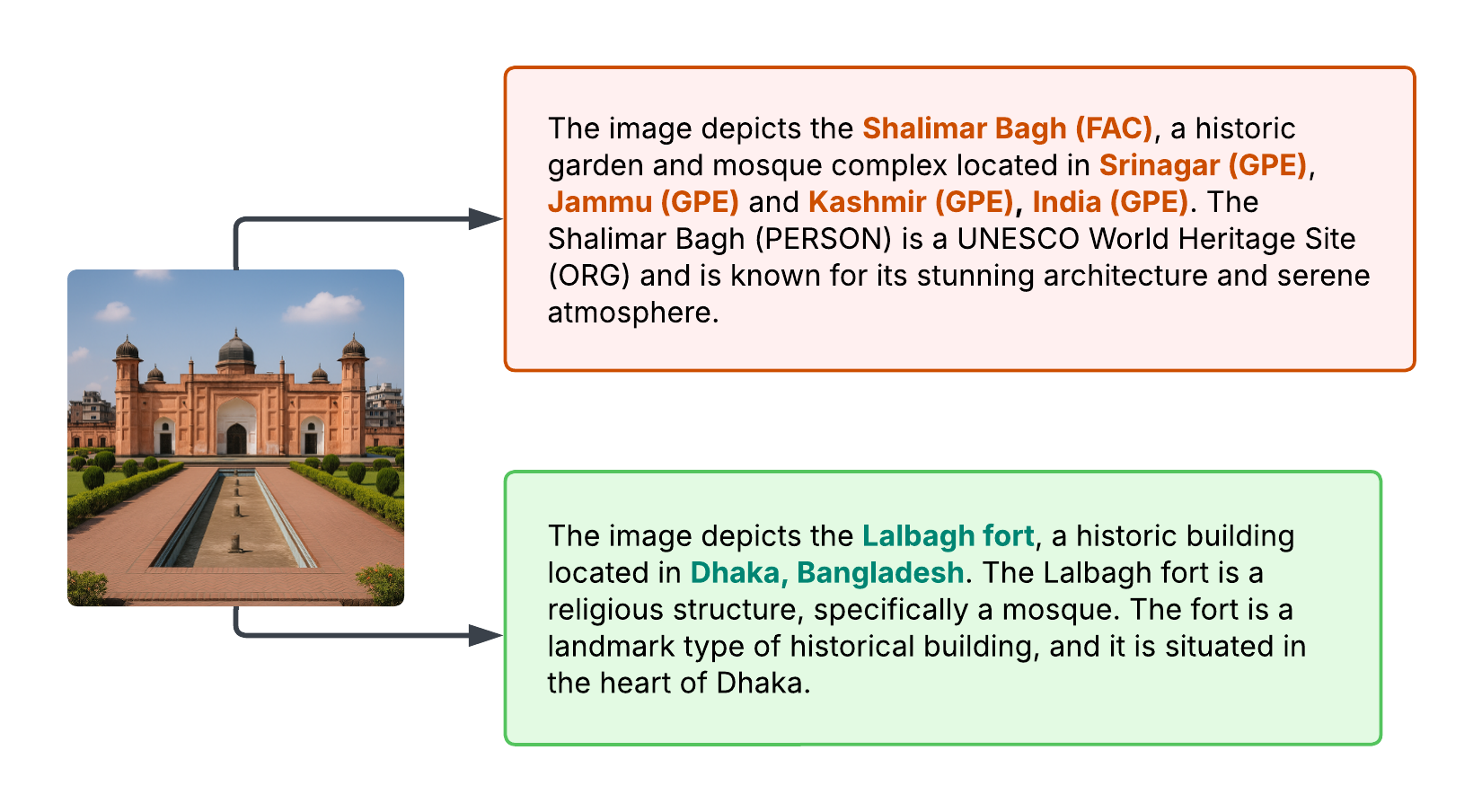}
  \caption{A comparison of hallucinated entities in red and the factually correct entities after processing through our pipeline in green.}
  \label{fig:side by side comparison}
\end{figure}
Visual Language Models have transformed image understanding tasks, yet their inability to reason systematically about facts within images and text remains a critical limitation. While humans naturally verify visual information against their knowledge base, VLMs lack structured mechanisms for fact verification, leading to confident but incorrect assertions about entities, locations, and relationships in images. This poses significant concerns about trustworthiness and reliability about their generated responses making them unreliable for domains like healthcare, education and cultural preservation. 

Unlike in LLMs, where integrating external knowledge for reasoning is actively studied using retrieval \citet{lewis2021retrievalaugmentedgenerationknowledgeintensivenlp} and in-context learning \citet{brown2020languagemodelsfewshotlearners}, systematic reasoning in VLMs remains underexplored particularly in tasks requiring fact verification and multi-step logical inference. This is especially problematic for factual verification tasks, where models must not only recognize visual elements but also reason about their relationships with real-world knowledge. Current VLMs lack structured mechanisms to perform such reasoning, often resulting in descriptions that combine accurate visual observations with incorrect factual assertions. For example, while a VLM might correctly identify architectural features of a historical landmark, it fails to systematically verify and reason about crucial facts like its location, historical significance, or cultural context. \\
This limitation stems from the absence of explicit reasoning paths between visual perception and knowledge integration, leading to unreliable factual claims and compromised utility in applications requiring high factual precision. VLMs face several unique technical challenges: they must jointly align facts with both image and textual data simultaneously, traverse complex knowledge structures across multiple reasoning steps, and maintain consistency between visual evidence and external knowledge sources. \\

These challenges are compounded by the need to represent knowledge in a format that supports both visual grounding and logical inference. To address these challenges, we propose a structured reasoning framework that explicitly models the verification path from visual perception to knowledge integration through multiple coordinated hops. To address these challenges, we introduce a multi-hop reasoning framework that enables VLMs to perform structured verification using knowledge graphs. \\

Our framework decomposes the verification process into distinct reasoning hops: entity recognition from visual inputs, knowledge graph traversal for fact retrieval, and structured verification of generated descriptions. Each hop is designed to maintain interpretable reasoning paths, allowing the model to explicitly track how it verifies facts against both visual evidence and knowledge sources. We implement this through three key innovations: (1) a hierarchical knowledge representation that supports both visual and factual reasoning, (2) a structured verification mechanism that traces reasoning paths through the knowledge graph, and (3) an adaptive correction strategy that resolves conflicts between visual observations and stored knowledge. Central to our approach is the flexible use of different knowledge representation formats namely hierarchical trees, relation triples, and structured facts each optimized for different types of reasoning tasks. This flexibility allows the model to choose appropriate reasoning paths based on the verification task, whether comparing spatial relationships, verifying historical facts, or checking entity attributes. We provide detailed ablation studies to trace the model's reasoning in each of these knowledge representation. Through extensive experimentation on landmark description tasks, we demonstrate that our framework significantly reduces hallucination while improving factual consistency in generated descriptions.



\section{Related Studies}
\subsection{Vision-Language Models for Image Captioning}
Vision-language models are now being for image captioning in the wild by leveraging their dual architectures to generate and align multi-modal embeddings. Early models, such as Show and Tell \citet{vinyals2015tellneuralimagecaption}, paired convolutional neural networks with recurrent neural networks for image description tasks. With the widespread development of transformer-based architectures, models like CLIP \citet{radford2021learningtransferablevisualmodels} demonstrated state-of-the-art performance by aligning textual and visual embeddings through contrastive learning. Recent work has further demonstrated that incorporating external knowledge can significantly improve vision-language models' factual accuracy and reasoning capabilities as shown by Anderson et al. \citet{anderson2018bottomuptopdownattentionimage}. Building on this, Zhang et al. \citet{kang2023knowledgegraphaugmentedlanguagemodels} proposed a knowledge-aware transformer architecture that explicitly reasons over both visual features and structured knowledge.
\subsection{Studying hallucinations in image captioning task}
Despite these advancements, VLMs often generate captions with hallucinated or factually inaccurate entities, especially for unseen or domain-specific data as supported by Rohrbach et al. \citet{rohrbach2019objecthallucinationimagecaptioning}. We were particularly inspired by their findings of how image captioning models often fail to capture image relevance with their internal understanding. Techniques such as entity-aware training \citet{cao2021cliff} have been proposed to mitigate this issue. Integrating structured knowledge, such as ontologies or databases, has shown promise in improving factual accuracy for image captioning tasks. For instance, memory-augmented models \citet{cornia2020meshedmemorytransformerimagecaptioning} that retrieved relevant facts to enhance descriptions were introduced to overcome this limitation. This kind of inaccuracies limit VLMs usage in applications requiring factual precision, such as education, historical archiving etc.
\subsection{Knowledge Graph Integration Approaches}
Different approaches for integrating knowledge graphs with neural models have been explored. There has been significant efforts to integrate KGs into neural models and also into use language models to enhance KGs such as KG-BERT \citet{yao2019kgbertbertknowledgegraph} and TransE \cite{Bordes2013TranslatingEF}, which embed graph knowledge into vector spaces for downstream tasks. In addition, Kumar et al. \citet{zhou2024knowledgeenhancedvisuallanguagepretrainingcomputational} studied projecting domain specific knowledge from a custom curated knowledge base into the latent space of the language model for pretraining the LM. 

The use of KGs remains underexplored for image-captioning tasks especially when we want to understand how VLMs jointly reason about the image and their underlying knowledge representation. Our work in integrating VLMs with KGs for factual image captioning, as explored in this study, builds upon these foundational works, addressing the gap in ensuring factual accuracy in multi-modal systems. Our contributions are as follows:
\begin{itemize}
\item  We propose a straight-forward integration of knowledge graphs illustrated with the example of landmark identification 

\item Our approach combines both pre-generation knowledge integration and post-generation correction 

\item We introduce new methods for comparing different knowledge representation formats

\item  We provide quantitative analysis of factual accuracy improvements

\end{itemize}

\section{Multi-Hop Reasoning Framework}

\begin{figure*}[t]
  \includegraphics[width=\linewidth]{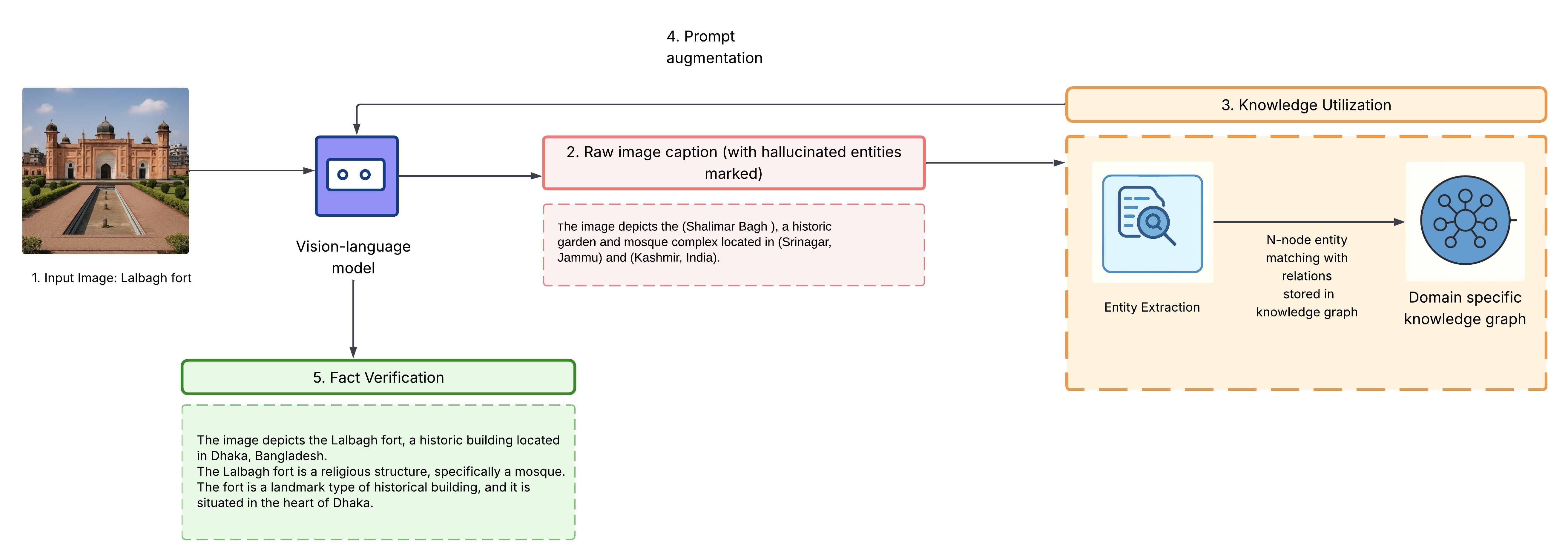} \hfill
  \caption {The system ingests an input image, generates a base caption using a VLM, and sequentially refines the caption through entity extraction, knowledge graph matching and augments the corrected entities to the prompt of the vlm to generate a factually accurate caption. Each module is color-coded and can operate independently, allowing for modular reasoning and analysis.}
\end{figure*}
We introduce a multi-hop reasoning framework that systematically verifies and corrects these descriptions using structured knowledge as shown in Figure 2. Our framework consists of five key components that progressively refine the caption generation process:

\textbf{Vision-Language Understanding} The initial component leverages a pre-trained VLM (Qwen2-VL-2B-Instruct) \citet{Qwen2VL} to generate a base caption C from input image I. While this produces fluent descriptions, we observed that approximately 69\% of entity mentions were either incorrect or hallucinated. 

\textbf{Entity Extraction Hop} This component extracts named entities E = {\( e_1 \),...,\( e_n \)} from C using spaCy's NER model \citet{spacy2}. We focus on entities across multiple categories including locations, organizations, and facilities to capture the full range of factual claims made in the caption. 

\textbf{Knowledge Graph Navigation} For each extracted entity \( e_i \), we perform both exact and fuzzy matching against our knowledge graph, G. Fuzzy matching uses sentence embeddings (all-MiniLM-L6-v2) to identify the closest entity in G when exact matches fail. This produces two sets: verified entities V and potentially hallucinated entities H. 

\textbf{Fact Verification} This stage validates the relationships between verified entities using three knowledge representation formats:
\textbf{Triple-based verification} - (subject, relation, object), 
\textbf{Hierarchical path validation} - ancestor-descendant relationships, 
\textbf{Bullet-point fact matching} key-value attribute pairs of entities in the raw caption.
The choice of multiple formats enables robust cross-validation while handling different types of factual claims. \\

\textbf{Caption Correction} Finally, we generate a corrected caption C' by integrating the verified facts with the original caption structure. This maintains the fluent language of the VLM while ensuring factual accuracy. The correction process uses prompt engineering to preserve proper context and coherence.

Each hop in our framework produces interpretable intermediate outputs, allowing for fine-grained analysis of the reasoning process. The modular design also enables easy integration of additional knowledge sources and reasoning strategies

\subsection{Exploring knowledge representation formats}
The choice of knowledge representation significantly impacts how factual information can be verified and integrated into image captions. While traditional knowledge graphs excel at capturing entity relationships, we found that different reasoning tasks benefit from complementary representation formats. Our framework utilizes three distinct knowledge representations, each offering unique advantages for fact verification and caption correction.
\subsubsection{Triple-based Relations}
The most simplest knowledge representation is a triple-based format that expresses facts as (subject, relation, object) statements. For example: \\
(Lalbagh fort, Located\_In, Dhaka),
(Dhaka, Capital\_Of, Bangladesh),
(Lalbagh fort, religious\_structure, mosque)

This representation excels at capturing direct relationships and enables efficient graph traversal for multi-hop reasoning. However, we observed that triples alone sometimes fail to capture hierarchical information and complex contextual relationships.

\subsubsection{Hierarchical Knowledge Trees}
To address the limitations of flat triples, we implement a hierarchical representation that captures nested relationships:
\begin{itemize}
    \item Lalbagh Fort
    \begin{itemize}
        \item Located In: Dhaka
        \begin{itemize}
          \item Capital Of: Bangladesh
        \end{itemize}
        \item Type: historical building
        \item Structure: mosque
    \end{itemize}
\end{itemize}

This format's key advantage is its ability to represent containment relationships and inheritance properties naturally. It particularly aids in correcting location-based errors in captions by providing clear geographical hierarchies. However, the hierarchical format can make some types of transitive reasoning more computationally expensive.

\subsubsection{Bullet-point Facts} 
We also explored a simplified representation in bulleted form that focuses on direct attribute-value pairs:
\begin{itemize}
    \item Lalbagh fort: Located In Dhaka
    \item Lalbagh fort: religious structure mosque
    \item Lalbagh fort: landmark type historical building
\end{itemize}

This format provides quick fact lookup and is especially effective for prompt engineering in the caption correction phase. Its simplicity makes it ideal for direct entity attribute verification, though it sacrifices the ability to perform complex reasoning.

Our flexible framework allows mixing and matching or using any of the frameworks in isolation. When verifying entities and relationships, we first consult the triple-based representation for explicit relationships, then use the hierarchical format for containment verification, and finally reference the bullet-point facts for direct attribute confirmation. This multi-representation approach significantly improves the robustness of our fact verification process, achieving a 27\% reduction in hallucinated entities compared to using triple-based representation alone.

\section{Evaluation}
We evaluate our knowledge-augmented reasoning framework through a series of experiments designed to answer three key research questions about knowledge representation, reasoning paths, and failure modes. We used the Qwen family of models. We systematically studied the Fact Verification Rate with each knowledge representation by keeping the model size constant. We preliminarily share detailed numbers for the 2 billion variant of the model.
\subsection{Experimental Setup}
 Our dataset combines images from Google Landmarks Dataset v2 (GLDv2) \citet{weyand2020GLDv2}, Conceptual Captions \citet{sharma2018conceptual}, and COCO Captions \citet{chen2015microsoftcococaptionsdata} to create a diverse, multi-domain collection for factual reasoning and knowledge graph-based verification. We select landmark-centric images from GLDv2 as ground truth, supplement with landmark-related images from Conceptual Captions for generalization challenges, and include COCO Captions samples to test robustness across everyday scenes. The dataset is partitioned into three splits to evaluate multi-hop reasoning and knowledge generalization: 1) Seen Landmarks (60\%) - entities and relationships present in the knowledge graph, 2) Unseen Landmarks (20\%) - landmarks absent from the knowledge graph but with related higher-level entities, and 3) Distractor Scenes (20\%) - non-landmark or ambiguous scenes testing hallucination detection and entity verification. This balanced split enables rigorous evaluation of both in-domain and out-of-domain reasoning while assessing the system's ability to reject hallucinated entities and perform multi-hop verification.

\subsubsection{Evaluation Metrics}
For our purpose, we had to come up with custom evaluation metrics that represent our use case. We detail the definitions and corresponding formulae to calculate them as well. 

\textbf{Entity Accuracy}: Percentage of correctly identified entities 
\begin{equation}
\label{eq:EA}
    EA = \frac{\text{NME} + \text{NHC}}{\text{NTE}} \times 100\%
\end{equation}

Where NME is the number of Matched Entities (entities correctly identified and matched to knowledge graph), NHC is the number of Hallucinations Correctly Identified (false entities properly detected) using threshold and NTE is the number of total entities mentioned in caption.

\textbf{Fact Verification Rate}: Proportion of successfully verified factual claims 
\begin{equation}
\label{eq:FVR}
    FVR = \frac{\text{NCV}}{\text{NTC}} \times 100\%
\end{equation}
Here, NCV is the number of Correctly Verified Facts and NTC is the number of Total Claims made in the caption.
Furthermore, NTC is calculated by the formula 
\begin{equation}
\label{eq:NTC}
    NTC = \text{NEC} + \text{NLC} + \text{NAC} + \text{NRC}
\end{equation}
We define NEC as the number of Entity Claims (basic existence claims), NLC as number of Location Claims (spatial relationships), NAC is the number of Attribute Claims (properties/characteristics) and NRC is the number of Relationship Claims (connections between entities). 

\textbf{Caption Coherence (Cc)}: Human evaluation of caption fluency (1-5 scale) \\

\begin{table}
  \centering
  \begin{tabular}{lc}
    \hline
    \textbf{Format} & \textbf{EA} \\
    \hline
    Triples Only & 72.3\%           \\
    Hierarchical Only & 78.1\%           \\
    Bullet-points Only & 65.7\%          \\\hline
  \end{tabular}
  \begin{tabular}{lc}
    \hline
    \textbf{FVR} & \textbf{Cc} \\
    \hline
    68.5\% & 4.2         \\
    73.2\% & 4.1          \\
    61.8\% & 4.3            \\
    \hline
  \end{tabular}
  \caption{Qwen-VL-2b analysis on different knowledge representations}
  \label{tab:accents}
\end{table}
We report a 31.8\% improvement in factual accuracy, measured as the reduction in hallucinated entities within generated captions after applying our multi-hop reasoning framework.
Hallucinated entities were identified based on: (1) absence in the domain-specific knowledge graph, (2) failure to meet the confidence threshold during entity matching (< 0.85), and (3) inability to verify factual claims through knowledge representation formats. All entities were manually annotated for reliability.
The relative improvement was calculated as:
\begin{equation}
\label{eq:factual_accuracy}
\text{FI (\%)} = \frac{ N_{H}^{\text{baseline}} - N_{H}^{\text{corrected}} }{ N_{H}^{\text{baseline}} } \times 100
\end{equation}

Here the Factual Improvement, FI is calculated as Number of Hallucinated Entities in Baseline Captions minus the Number of Hallucinated Entities after Correction.

On our evaluation set of 100 images, baseline VLM captions contained 55 hallucinated entities versus 38 in corrected captions, yielding 31.8\% improvement. This demonstrates our system's effectiveness in reducing factual hallucinations through systematic multi-hop reasoning and knowledge-guided correction.
Our key findings are tabulated in Table 1 and we highlight the following observations:
\begin{itemize}
    \item Hierarchical representation performs best in isolation, particularly for spatial reasoning related image understanding. However this format limits the model's free-format generation abilities leading to a decrease in caption coherence.
    \item Bullet-point format, while simplest, maintains highest caption coherence, we hypothesize that this maybe due to under
\end{itemize} 
\section*{Conclusion}

We presented a modular multi-hop reasoning system for improving factual accuracy in vision-language models through structured knowledge integration. The system enables step-by-step fact verification and caption correction by leveraging knowledge graphs and interpretable reasoning paths. Our prototype demonstrates an approximately 31.8\% reduction in hallucinated entities and highlights the potential of modular fact-checking pipelines in improving the reliability of multimodal systems. This work lays the foundation for scalable, knowledge-grounded captioning systems with applications in education, cultural heritage, and safety-critical domains.
\section*{Limitations}

This system is currently evaluated as a prototype on a small, domain-specific dataset of landmark images. While the modular multi-hop framework is generalizable, the current knowledge graph is manually curated and not yet scaled for open-domain applications. Additionally, the system is not currently evaluated on other VLMs for image captioning. Future work will focus on expanding dataset coverage, integrating large-scale dynamic knowledge sources, and improving evaluating multimodal models.

\bibliography{custom}

\appendix
\section{Appendix}
\label{sec:appendix}
\subsection{Relevant Knowledge Graph construction}
The effectiveness of knowledge-augmented image captioning substantially depends on the quality and coverage of the underlying knowledge graph. We describe our systematic approach to constructing a focused yet comprehensive knowledge graph for validating image descriptions that can be adapted for any custom use case. 
\subsubsection{Entity Selection}
We adopt a domain-driven approach to entity selection to illustrate our purpose. As we focused on architectural landmarks and their geographical contexts we provide some example criteria for entity selection approach for this problem below.
\begin{itemize}
    \item Architectural significance: Historical buildings, monuments, and religious structures that are frequently referenced in image descriptions
    \item Geographical hierarchy: Administrative divisions, cities, and countries that provide crucial location context
    \item Architectural attributes: Physical characteristics and historical features that help validate visual descriptions
\end{itemize}
This targeted selection enables efficient verification while maintaining high coverage of relevant entities. For example, in our implementation focused on the cultural heritage sites such as the Lalbagh Fort that are not commonly represented in openly available dataset and thus provide an excellent data point to study the behaviour of VLMs while genrating captions. 

\subsubsection{Relation Definition}
Relations in our knowledge graph fall into three categories, each serving a specific verification purpose:
Spatial Relations e.g Located\_In: Captures physical containment (e.g., fort located in city), Capital\_Of: Represents administrative relationships 

Structural Relations e.g religious\_structure links buildings to their architectural type and landmark\_type which defines the category of historical structures

We carefully constrain relation definitions to maintain consistency and enable reliable reasoning. Each relation type is explicitly typed and directional, facilitating both forward and backward traversal so that we can also trace the vlms reasoning trajectory for verification. 

\subsubsection{Graph Connectivity}

The connectivity of our knowledge graph is designed to support multi-hop reasoning while avoiding spurious connections. We implement this through: hierarchical connections \\

\begin{algorithm}
   \label{alg: hierarchical}
\begin{algorithmic}
    \STATE $G.add\_edge(Lalbagh fort, Dhaka, relation=located\_in)$
    \STATE $G.add\_edge(Dhaka, Bangladesh,$
    \STATE \quad $relation=captital\_of)$
\end{algorithmic}
\end{algorithm}

and attribute connections  \\

\begin{algorithm}
   \label{alg:attribute}
\begin{algorithmic}
    \STATE $G.add\_edge(Lalbagh fort, mosque, relation=religious\_structure)$
    \STATE $G.add\_edge(Lalbagh fort, historical\_building,$
    \STATE \quad $relation=landmark\_type)$
\end{algorithmic}
\end{algorithm} 
The resulting graph exhibits several desirable properties:
\begin{itemize}
    \item Average node degree: 2.5, ensuring sufficient connectivity for reasoning
    \item \(Path\_length \leq 3 \) between any related entities, enabling efficient verification
    \item Hierarchical clustering coefficient: 0.67, reflecting strong local structure
\end{itemize}
This structured approach to knowledge graph construction provides a strong foundation for our fact verification system, while the careful curation of entities and relations helps minimize computational overhead during the verification process making systematic experimentation easier. 
\subsection{Reasoning Pipeline}
Our system implements a modular reasoning pipeline that systematically processes image captions through multiple stages. The pipeline architecture follows a hop-based design pattern, where each hop represents a discrete reasoning step with well-defined inputs and outputs.

Each hop in the pipeline serves a specific reasoning function:
\begin{enumerate}
    \item Entity Recognition Hop: Processes raw captions using spaCy's NER model to identify named entities, locations, and key architectural terms. The hop outputs a structured list of entities:
   
    Entities = \{ \\
                    FAC: ['the Lalbagh fort'], \\
                    GPE: ['Dhaka', 'Bangladesh'], \\
                    ORG: ['UNESCO World Heritage Site'] \\
\}
   
    \item Knowledge Graph Hop: Maps identified entities to our knowledge graph through both exact and fuzzy matching. The hop employs sentence embeddings (all-MiniLM-L6-v2) for fuzzy matching, producing two entity sets: 1) Verified-entities: Entities matched with \(confidence score \geq 0.85 \), 2) Potential hallucinations: Unmatched or low-confidence matches.
    \item Verification Hop: Validates relationships between verified entities using our multi-format knowledge representation. The hop generates a verification report containing 1) Confirmed facts 2) Identified discrepancies and 3) Confidence scores for each verification. 
    
    \item Correction Hop: Synthesizes the verification results to produce a factually accurate caption while maintaining natural language fluency. This hop employs template-based correction strategies depending on the type of factual error identified.
Our implementation ensures each hop operates independently while maintaining a cohesive reasoning chain. The modular design allows for easy integration of new reasoning components
and parallel processing where applicable.

\end{enumerate}



\end{document}